\documentclass{isprs} 
\usepackage{subfigure}
\usepackage{setspace}
\usepackage{geometry} 
\usepackage{epstopdf}
\usepackage[labelsep=period]{caption}  
\usepackage[british]{babel} 
\usepackage[hang]{footmisc}

\usepackage[authoryear,round,longnamesfirst]{natbib}
\usepackage[acronym]{glossaries}
\newacronym{ALS}{ALS}{airborne laser scanning}
\newacronym{MLS}{MLS}{mobile laser scanning}
\newacronym{MLS2}{MLS}{mobile laser scanners}
\newacronym{LoD}{LoD}{level of detail}
\newacronym{LoDv2}{LoD}{levels of detail}
\newacronym{OGC}{OGC}{open geospatial consortium}
\newacronym{GML}{GML}{geography markup language}
\newacronym{ASAM}{ASAM}{association for standardization of automation and measuring systems}
\newacronym{TLS}{TLS}{terrestrial laser scanning}
\newacronym{UAV}{UAV}{unmanned aerial vehicle}
\newacronym{HD}{HD}{high definition}
\newacronym{RANSAC}{RANSAC}{random sample consensus}
\newacronym{ROI}{ROI}{region of interest}
\newacronym{DEM}{DEM}{digital elevation model}
\newacronym{ICP}{ICP}{iterative closest point}
\newacronym{NLOS}{NLOS}{non-line-of-sight}
\newacronym{SfM}{SfM}{structure from motion}
\newacronym{FME}{FME}{feature manipulation engine}
\newacronym{OSM}{OSM}{OpenStreetMap} 
\newacronym{RMSE}{RMSE}{root mean square error}
\newacronym{CPT}{CPT}{conditional probability table}
\newacronym{DST}{DST}{Dempster–Shafer theory}
\newacronym{BN}{BayNet}{Bayesian network}
\newacronym{GIS}{GIS}{geographic information system}
\newacronym{PPD}{PPD}{posterior probability distribution}
\newacronym{CI}{CI}{confidence interval}
\newacronym{CL}{CL}{confidence level}
\newacronym{LiDAR}{LiDAR}{light detection and ranging}
\newacronym{CSG}{CSG}{constructive solid geometry}
\newacronym{BIM}{BIM}{building information modeling}
\newacronym{TUM}{TUM}{the Technical University of Munich}
\newacronym{CRS}{CRS}{coordinate reference system}
\newacronym{MVS}{MVS}{multi-view stereo}
\newacronym{GNSS}{GNSS}{global navigation satellite system}

\begin{document}

\title{Combining visibility analysis and deep learning for refinement of semantic 3D building models by conflict classification}

\author{
 O. Wysocki\textsuperscript{1}\thanks{Corresponding author}, E. Grilli\textsuperscript{3}, L. Hoegner\textsuperscript{1,2}, U. Stilla\textsuperscript{1}
}

\address{
	\textsuperscript{1 }Photogrammetry and Remote Sensing, TUM School of Engineering and Design, Technical University of Munich (TUM),\\ Munich, Germany - (olaf.wysocki, ludwig.hoegner, stilla)@tum.de\\
	\textsuperscript{2 }Department of Geoinformatics, University of Applied Science (HM), Munich, Germany - ludwig.hoegner@hm.edu\\
	\textsuperscript{3 }3D Optical Metrology (3DOM) unit, Bruno Kessler Foundation (FBK), Trento, Italy - grilli@fbk.eu
}


\commission{}{} 
\workinggroup{} 
\icwg{}   

\abstract{

Semantic 3D building models are widely available and used in numerous applications.
Such 3D building models display rich semantics but no façade openings, chiefly owing to their aerial acquisition techniques.
Hence, refining models' façades using dense, street-level, terrestrial point clouds seems a promising strategy.
In this paper, we propose a method of combining visibility analysis and neural networks for enriching 3D models with window and door features.
In the method, occupancy voxels are fused with classified point clouds, which provides semantics to voxels.
Voxels are also used to identify conflicts between laser observations and 3D models.
The semantic voxels and conflicts are combined in a Bayesian network to classify and delineate façade openings, which are reconstructed using a 3D model library.
Unaffected building semantics is preserved while the updated one is added, thereby upgrading the building model to LoD3.
Moreover, Bayesian network results are back-projected onto point clouds to improve points' classification accuracy.
We tested our method on a municipal CityGML LoD2 repository and the open point cloud datasets: TUM-MLS-2016 and TUM-FAÇADE.
Validation results revealed that the method improves the accuracy of point cloud semantic segmentation and upgrades buildings with façade elements.
The method can be applied to enhance the accuracy of urban simulations and facilitate the development of semantic segmentation algorithms.

}

\keywords{3D reconstruction, MLS point clouds, Semantic 3D building models, CityGML, Deep learning, LoD3 building models, Window and door reconstruction, Building models refinement.}
\maketitle

\section{Introduction}\label{sec:Introduction}
\sloppy

\begin{figure} [htb]
    \centering
    \includegraphics[width=\linewidth]{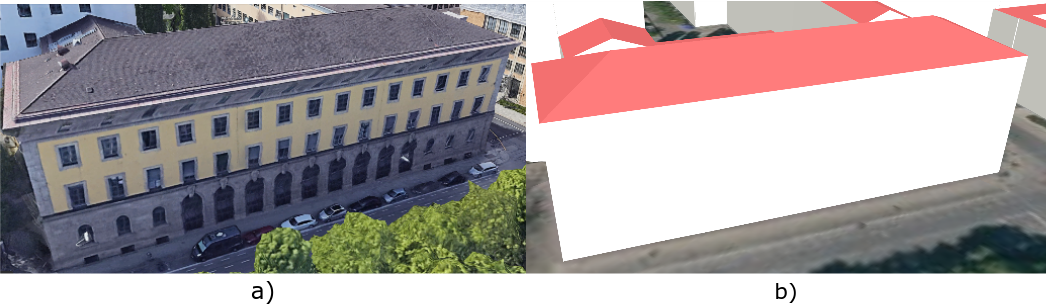}
    \caption{Façade in a photo and a 3D building model: a) Oblique image~\citep{googleEarthfrontPage}, b) semantic building model at LoD2~\citep{bayernAtlasfrontPage}.}
    \label{fig:frontFigure}
\end{figure}

Semantic 3D building models at \gls{LoDv2}1 and 2 are widespread~\footnote{https://github.com/OloOcki/awesome-citygml} and commonly applied in urban-related studies~\citep{biljeckiApplications3DCity2015}.
Such 3D models are frequently reconstructed using a combination of 2D building footprints and~\gls{MVS} or~\gls{ALS} techniques, as in the example of more than eight million reconstructed buildings in Bavaria, Germany~\citep{RoschlaubBatscheider}.
This reconstruction strategy enables detailed modeling of roof surfaces but renders generalized façades neglecting openings such as windows and doors, as shown in Figure~\ref{fig:frontFigure}b. 

Reconstructing façade elements becomes a key factor enabling automatic 3D building modeling at \gls{LoDv2}3, for which an increasing demand has been expressed by numerous applications including estimating heating demand~\citep{nouvel2013citygml}, preserving cultural heritage~\citep{Grilli2019}, calculating solar potential~\citep{willenborgIntegration2018}, and testing automated driving functions~\citep{schwabRequirementAnalysis3d2019}. 

Since point clouds are deemed as one of the best data sources for 3D modeling, dense, street-level~\gls{MLS} point clouds appear to be especially suitable for at-scale façade reconstruction~\citep{xu2021towards}.
For this purpose, however, point clouds require semantic classification, which has been recently approached using machine and deep learning methods yielding promising results~\citep{grilli2020machine,matrone2020comparing}. 
Yet, these methods can have limited accuracy when classifying objects that are translucent (e.g., windows) or have an inadequate amount of training data (e.g., doors).
On the other hand, points' rays intersecting with 3D models can provide geometrical cues about possible façade openings, but without differentiating between classes, such as window, door, or underpass~\citep{tuttas_reconstruction_2013}. 

In this paper, we present a strategy that combines both ray- and region-based methods for conflict classification~\citep{wysockiUnderpasses}.
This approach leads to refinement of both 3D building models and segmented point clouds' accuracy; our contributions are as follows:
\begin{itemize}
    \item a CityGML-compliant strategy for upgrading \gls{LoDv2}2 to \gls{LoDv2}3 models by model-driven 3D window and door reconstruction; 
    \item a method classifying conflicts between laser observations and 3D building models using deep learning networks; 
    \item a method improving the semantic segmentation results of deep learning networks by analyzing ray-traced points and 3D building models.
\end{itemize}

\section{Related work}\label{sec:RelatedWorks}

The internationally used CityGML standard establishes the \gls{LoDv2} of semantic 3D city objects~\citep{grogerOGCCityGeography2012}.
One of the chief differences between \gls{LoDv2}2 and \gls{LoDv2}3 is the presence of façade openings in the latter.
In our case, we search for absent façade elements in the input models using point clouds and then carry out their reconstruction. 
Therefore, we deem methods as related if they deal with detecting missing features (Section~\ref{sec:visibilityRelated}) and façade reconstruction using point clouds (Section~\ref{sec:facdeSegmentationRelated}).

\subsection{Visibility analysis using point clouds}
\label{sec:visibilityRelated}

\citet{hebel2013change} employ visibility analysis to detect changes between different point cloud epochs.
The method addresses the uncertainty of \gls{ALS} measurements using the \gls{DST}.
Ray tracing on a voxel grid is introduced to identify \textit{occupied}, \textit{empty}, and \textit{unknown} states per epoch.
Based on the epochs comparison, they distinguish: \textit{consistent}, \textit{disappeared}, and \textit{appeared} states. 

Visibility analysis is utilized to remove dynamic objects from point clouds, too~\citep{gehrung2017approach}.  
\gls{MLS} observations' rays are traced on an efficient octree grid structure introduced by \citet{hornung2013octomap}.
Each traced ray provides occupancy probabilities, which are accumulated per voxel using the Bayesian approach. 
Moving objects are removed based on decreasing occupancy probability of ray-traversed voxels. 

A multimodal approach to visibility analysis is proposed by \citet{tuttas2015validation}.
They investigate how to monitor the progress of a construction site using photogrammetric point clouds and \gls{BIM} models.
The Bayesian approach and the octree grid structure is employed to analyze the points' rays and vector models. 
The as-is (point cloud) to as-planned (3D model) comparison differentiates between \textit{potentially built}, \textit{not visible}, and \textit{not built} model parts.

In our previous work~\citep{wysockiUnderpasses}, we introduced visibility analysis to refine semantic 3D building models with underpasses using \gls{MLS} point clouds.
The method compares ray-traced points with building objects on an octree grid in a probabilistic fashion.
Contours of underpasses are identified based on an analysis of conflicts between laser observations and building models, supported by vector road features.

\subsection{Façade openings reconstruction using point clouds}
\label{sec:facdeSegmentationRelated}

Substantial research effort has been devoted to methods using images for façade segmentation~\citep{szeliski2010computer}. 
Nevertheless, 2D images require additional processing to enable semantic 3D reconstruction. 
3D point clouds, however, provide an immediate 3D environment representation, which makes them one of the best datasets for urban mapping~\citep{xu2021towards}.

When analyzing laser observations, openings are often assumed to represent holes due to their translucent characteristic or face-intruded position ~\citep{tuttas_reconstruction_2013,fan2021layout}.
For example, windows are detected based on building interior points, which imply opening existence~\citep{tuttas_reconstruction_2013}. 
Borders of openings are delineated based on the ray tracing of interior points and the detected façade plane in point clouds.

\citet{zolanvari2018three} propose a slicing method to identify openings using horizontal or vertical cross-sections.
The method finds façade planes using the RANSAC algorithm and removes noisy points based on their deviations from the planes. 
Gaps occurring in horizontal or vertical cross-sections delineate possible openings. 

Layout graphs are proposed by 
\citet{fan2021layout} to identify façade structures.
Spatial relations among detected objects are encoded and exploited by the Bayesian framework to deduce the whole façade layout. 

Recently, however, data-driven methods based on machine and deep learning approaches have provided promising results for classifying point clouds, especially when using the self-attention mechanism~\citep{zhao2021point}.
These great strides have influenced façade segmentation of point clouds, too~\citep{grilli2020machine,matrone2020comparing}.
Modified versions of the DGCNN deep learning architecture are proposed to classify façade elements in point clouds~\citep{grilli2020machine}.
The method employs features stemming from machine learning approaches to improve deep learning network accuracy.

Little research attention has been given to investigating the automatic upgrade of \gls{LoDv2}2 to \gls{LoDv2}3 building models using point clouds, except, to the best of our knowledge, our previous works refining overall façade geometry~\citep{wysocki2021plastic,wysocki2021unlocking} and reconstructing underpasses~\citep{wysockiUnderpasses}.
However, related work is proposed by~\citet{KadaFacades} for detecting and reconstructing openings, not by point clouds but by exploiting the textures of semantic city models.
They apply the Faster R-CNN deep neural network to identify the bounding boxes of windows and doors on textured CityGML building models.
To minimize inaccuracies in the alignment of openings, they apply mixed-integer linear programming.
Then, bounding boxes serve as reconstructed opening elements in \gls{LoDv2}3 building models.

\section{Methodology} \label{sec:methodology}

In contrast to our previous work devoted to refining building models with underpasses~\citep{wysockiUnderpasses}, in this paper we focus on detecting and reconstructing outstanding façade openings, such as windows and doors.  
Moreover, our method refines point cloud segmentation by back-projecting classified conflicts onto the input point clouds.
\begin{figure} [htb]
    \centering
    \includegraphics[width=0.7\linewidth]{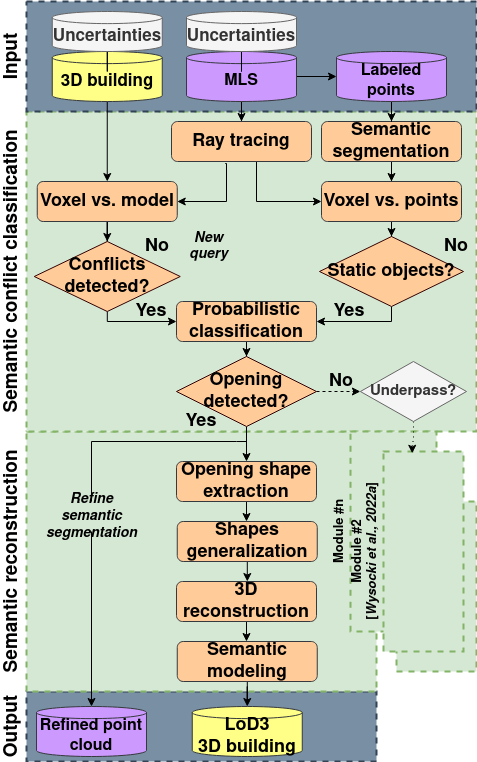}
    \caption{Workflow of the presented method}
    \label{fig:overviewModule}
\end{figure}

As presented in Figure~\ref{fig:overviewModule}, the method evaluates and assigns uncertainties to the input datasets (Section~\ref{sec:uncertainties}).
While a neural network is trained on points representing façade elements (Section~\ref{sec:semanticSegmentation}), the points ray tracing process performs probabilistic classification of a scene into \textit{occupied}, \textit{empty}, and \textit{unknown} voxels (Section~\ref{sec:rayCasting}).
Subsequently, labeled voxels are compared to segmented points to derive \textit{static} and remove \textit{dynamic} points in voxels (Section~\ref{sec:voxelToPoint}).
The voxels are also compared to vector 3D models to identify \textit{confirmed}, \textit{empty}, and \textit{unknown} voxel labels (Section~\ref{sec:voxelToModel}).
If \textit{conflicted} and \textit{static} features exist, probabilistic classification is carried out, where a Bayesian network identifies \textit{unmodeled openings} and \textit{other objects} (Section~\ref{sec:bayesian}).
These are back-projected to the point cloud, refining its segmentation accuracy.
If the Bayesian network detects windows or doors, shape extraction is conducted (Section~\ref{sec:shapeExtraction}); otherwise, another module can be triggered, such as the underpass reconstruction~\citep{wysockiUnderpasses}.
Opening shape extraction is followed by shape generalization, which delineates fitting borders for 3D reconstruction (Section~\ref{sec:shapeExtraction}).
Window and door 3D models are automatically fitted to shapes based on the respective geometry and opening class (Section~\ref{sec:libraryBased}).
Afterward, unchanged and new semantics are assigned to geometries, following the CityGML standard for \gls{LoDv2}3~\citep{grogerOGCCityGeography2012}.

\subsection{Data with uncertainties}
\label{sec:uncertainties}

Uncertainties in laser measurements and vector objects can stem from various sources, such as imprecise metadata, data transformations, and acquisition techniques.
Uncertainties are application-dependent, too.
Therefore, the proposed façades refinement involves uncertainties concerning the global positioning accuracy of point clouds and building models.
To quantify these uncertainties, we introduce the~\gls{CI}, which is estimated using the~\gls{CL}, its associated z value ($z$), standard deviation ($\sigma$), and mean ($\mu$).

Let~$\sigma_{1}$ be the location uncertainty of point clouds, and $\sigma_{2}$ the location uncertainty of 3D model walls.
These are estimated based on the assumed point cloud global registration error $e_{1}$ and the global location error of 3D model walls $e_{2}$.
Then, the façade's~\gls{CI} is calculated based on~$\sigma = \sqrt{\sigma_{1}^2 + \sigma_{2}^2}$.
The maximum upper and lower bounds are given by $[\mu_i  -  2\sigma_i, \mu_i  +  2\sigma_i]$, when assuming operating in the L1 norm and Gaussian distribution~\citep{suveg20003d}.
$CL_{1}$ and $CL_{2}$ quantify the operator's confidence level in true-value deviations for laser measurements and 3D model walls, respectively.
Depending on the~\gls{CL} value, corresponding $z_{i}$ values are assumed.
The division of $\mu_{i}$ by $z_{i}$ estimates the standard deviation $\sigma_{i}$ value~\citep{usingconfidenceZ}.

\subsection{Semantic segmentation} \label{sec:semanticSegmentation}

\begin{figure} [htb]
    \centering
    \includegraphics[width=0.9\linewidth]{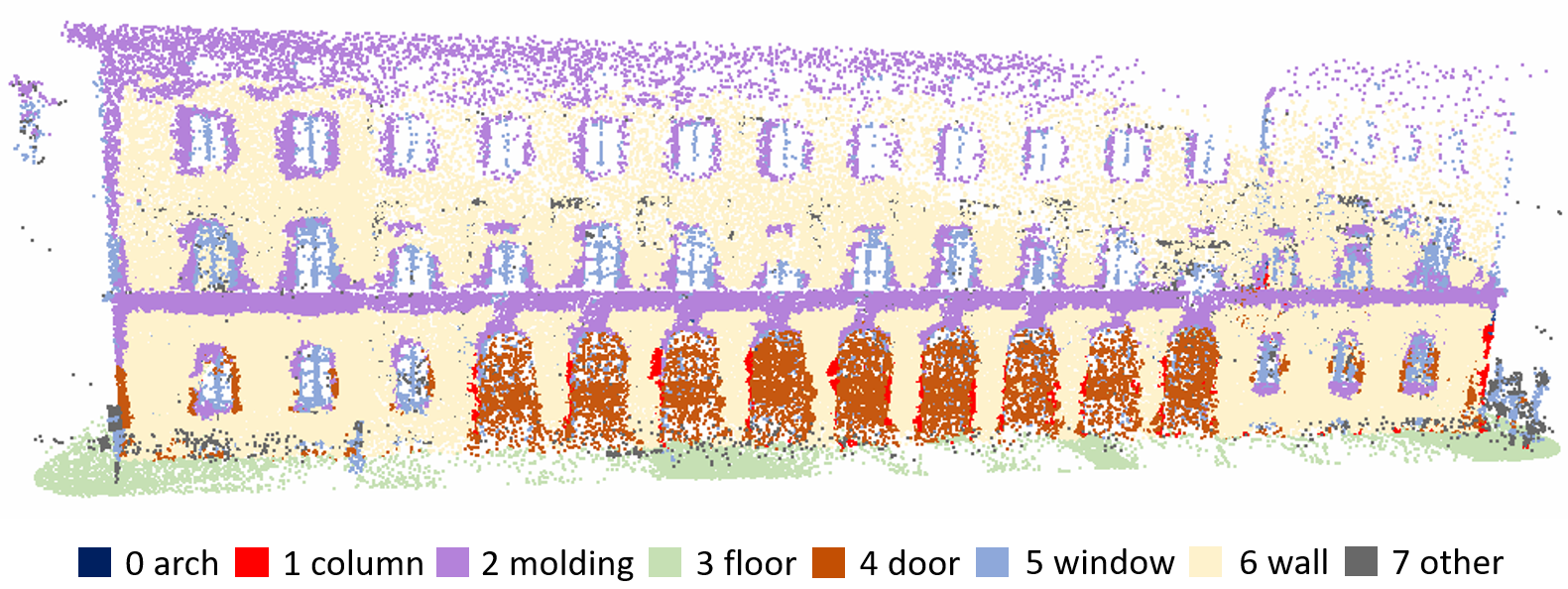}
    \caption{Semantic segmentation result for the façade in Figure~\ref{fig:frontFigure}.}
    \label{fig:semanticSeg}
\end{figure}

The goal of semantic segmentation is to divide a point cloud into several subsets based on the semantics of the points. 
Following \citet{tumfacadePaper} and as shown in Figure~\ref{fig:semanticSeg}, eight relevant classes for façade segmentation and reconstruction tasks are considered: \textit{arch} (dark blue), \textit{column} (red), \textit{molding} (purple), \textit{floor} (green), \textit{door} (brown), \textit{window} (blue), \textit{wall} (beige), and \textit{other} (gray). 

The segmentation is performed using a modified Point Transformer self-attention network~\citep{zhao2021point} extended by the use of geometric features improving the network performance, such as \textit{height of the points, roughness, volume density, verticality, omnivariance, planarity}, and \textit{surface variation}. 
The last three mentioned features are based on the normalized eigenvalues $\lambda_{i}$ $(\lambda_{1} > \lambda_{2} > \lambda_{3})$, which are derived from the 3D point coordinates within a considered spherical neighborhood $r_{i}$~\citep{jutziFeatures,grilli2020machine}. 

Finally, using a softmax output layer, we obtain an output vector of probabilities for each predicted class, which becomes fundamental for running our conflict classification approach (Section~\ref{sec:voxelToPoint}).

\subsection{Ray tracing} \label{sec:rayCasting}

\begin{figure}
    \centering
    \includegraphics[width=0.9\linewidth]{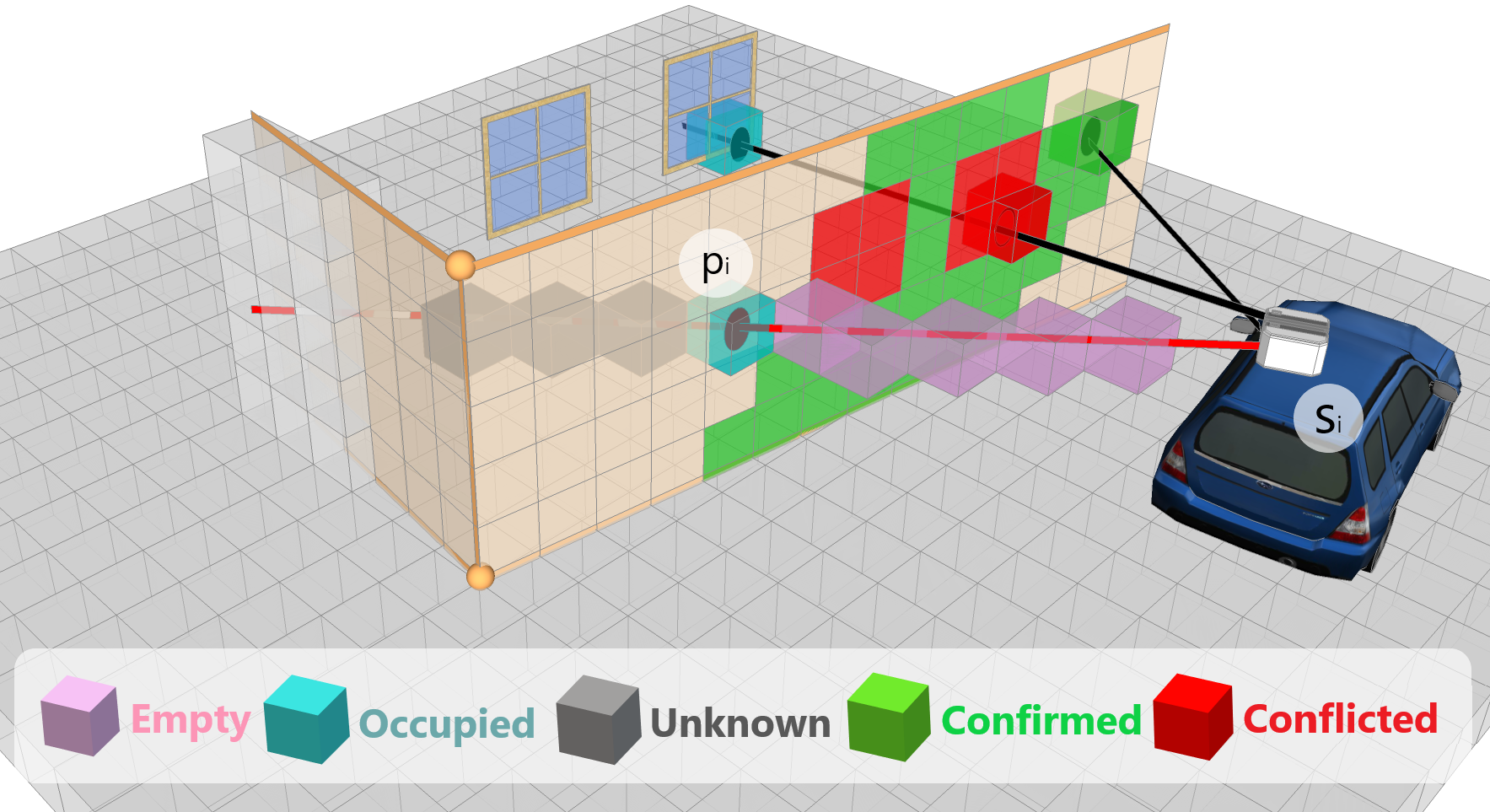}
    \caption{Points ray tracing on a vector-populated octree grid from the sensor position $s_{i}$ to the hit point $p_{i}$. Adapted from~\citep{wysockiUnderpasses}. }
    \label{fig:rayCasting}
\end{figure}

Points ray tracing is performed to identify absent structures in existing 3D building models (Figure~\ref{fig:rayCasting}).
To enable comparison between these modalities, we employ a 3D occupancy grid.
The grid adapts its size to the input data since it utilizes an octree structure.
3D voxels are the octree structure's leaves, and their size $v_{s}$ is selected based on the relative accuracy of laser observations. 

Every laser observation is traced from the sensor position $s_{i}$, following the orientation vector $r_{i}$, to the reflecting point $p_{i} = s_{i} + r_{i}$.
Voxels containing $p_{i}$ are labeled as \textit{occupied} (blue), those traversed by a ray as \textit{empty} (pink), and the untraversed ones as \textit{unknown} (gray).
The labels are assigned based on a probability score that considers multiple laser observations~$z_{i}$, which are updated using prior probability $P_(n)$ and previous estimate $L(n|z_{1:i - 1})$.
Final score is controlled using log-odd values~$L(n)$ and clamping thresholds $l_{min}$ and $l_{max}$~\citep{hornung2013octomap,tuttas2015validation}:
\begin{equation}
    L(n|z_{1:i}) = max(min(L(n|z_{1:i - 1}) + L(n|z_{i}), l_{max}),l_{min}) 
\end{equation}
where
\begin{equation}
     L(n) = log [\frac{P_{n}}{1 - P(n)}] 
\end{equation}

The grid is vector-populated by inserting 3D model faces and their quantified uncertainties (Section~\ref{sec:uncertainties}).
Hence, each face has an assigned façade's maximal deviation range (upper~\gls{CI}) and its confidence level (\gls{CL}).   
Ultimately, the grid's 3D voxels include attributes such as location, size, as well as state probability stemming from laser observations and a building model.

\subsection{Voxels to model comparison} \label{sec:voxelToModel}

As shown in Figure~\ref{fig:rayCasting}, each voxel is analyzed in relation to its intersection with a façade: 
\textit{Occupied} voxels that intersect with façades are labeled as \textit{confirmed} (green); \textit{empty} voxels that intersect with façades are labeled as \textit{conflicted} (red);
\textit{unknown} voxels hold their status, as they represent unmeasured space.
\begin{figure} [htb]
    \centering
    \includegraphics[width=0.7\linewidth]{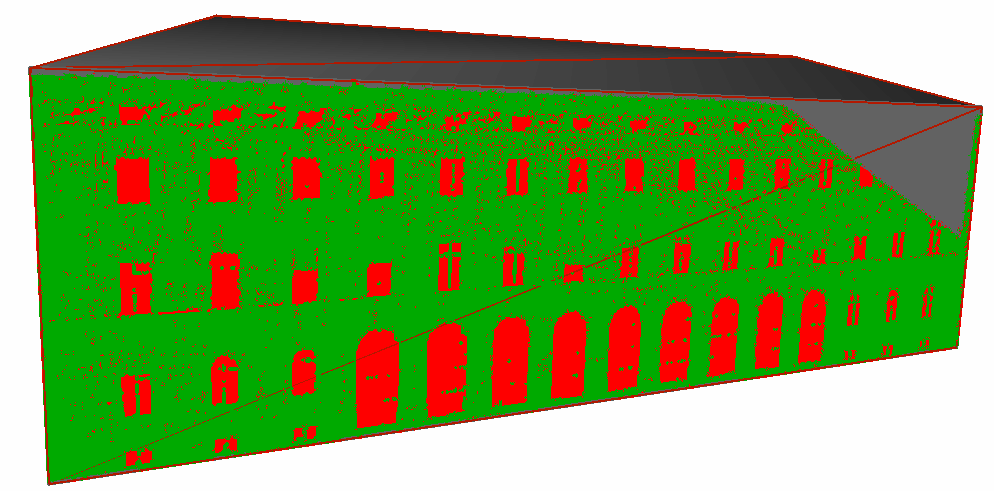}
    \caption{Texture representing \textit{confirmed}, \textit{conflicted}, and \textit{unknown} areas identified on a façade}
    \label{fig:textureConflicts}
\end{figure}
Voxels are projected onto the intersected façade, forming the \textit{model comparison} texture map layer with the respective voxel labels: \textit{confirmed}, \textit{conflicted}, and \textit{unknown} (Figure~\ref{fig:textureConflicts}).
The cell spacing of a texture map follows the projection of the voxel grid to the plane.


\subsection{Voxels to point cloud comparison} \label{sec:voxelToPoint}

\begin{figure} [htb]
    \centering
    \includegraphics[width=\linewidth]{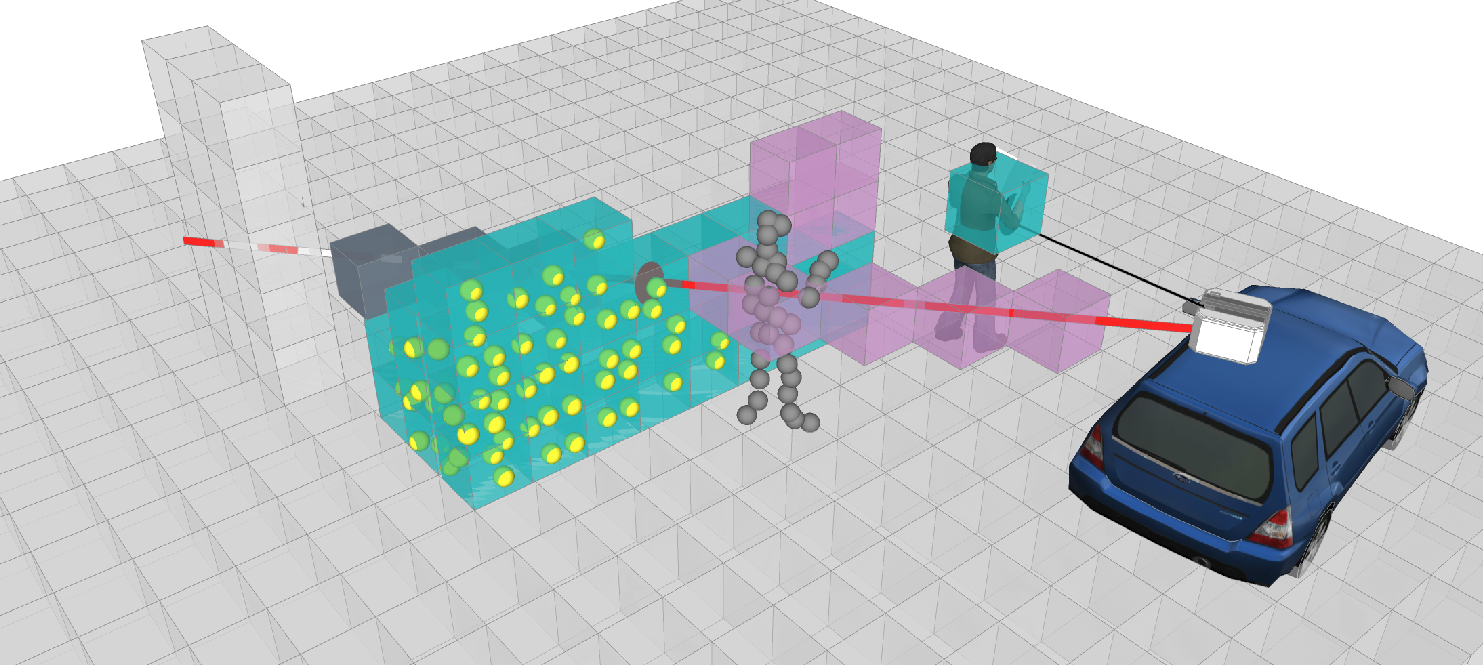}
    \caption{Fusion of voxels with per-point semantic information (yellow), while suppressing dynamic points (gray) using probability score and measurements accumulation.}
    \label{fig:fusionSemantic}
\end{figure}

\begin{figure} [htb]
    \centering
    \includegraphics[width=0.7\linewidth]{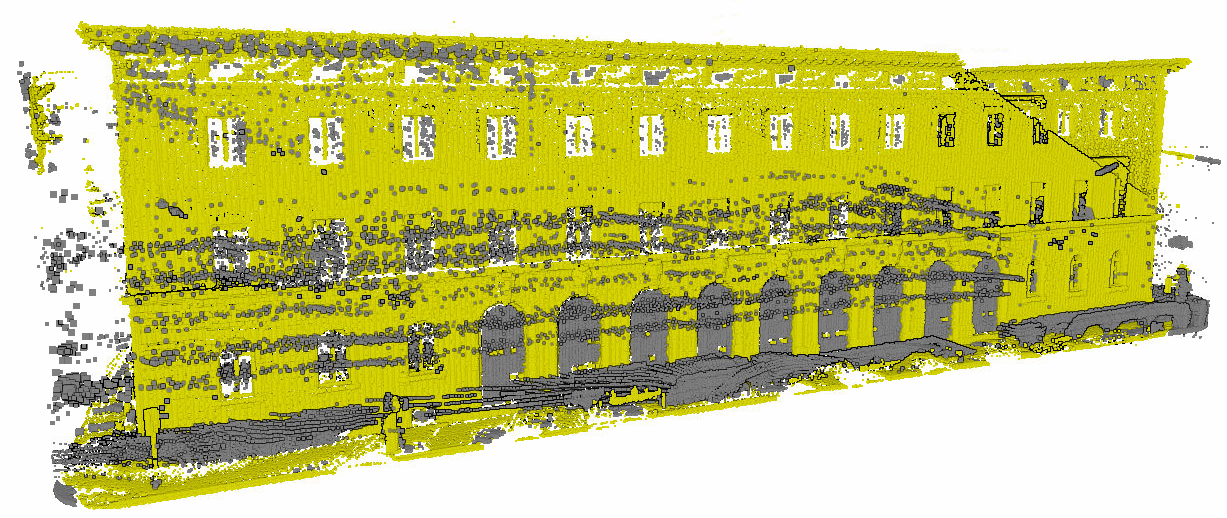}
    \caption{\textit{Dynamic}, noisy points (gray) separated from \textit{static}, building-related points (yellow).}
    \label{fig:dynamicVsStatic}
\end{figure}

Ray tracing provides physical, per-voxel occupancy indicators, while semantic segmentation yields educated, per-point semantic classes.
Both of these sources provide their semantic information with a probability measure.
The fusion of voxels and points is conducted to transfer per-point semantic classes to occupancy voxels and suppress the impact of dynamic points (Figure~\ref{fig:fusionSemantic}).
The rationale behind this fusion is that \textit{static}, occupied voxels (yellow) are building-related;
\textit{dynamic}, unoccupied voxels (gray) represent moving objects, such as pedestrians or cars, and can be suppressed by multiple laser observations, as shown by \citet{gehrung2017approach} and in Figure~\ref{fig:dynamicVsStatic}. 

Semantic points are inserted into the voxel grid to enable comparison between the two representations.
Then, the median probability score $P(B)$ is derived from point classes within each voxel.
The occupancy probability $P(A)$ and the median probability of each class $P(B)$ are two dependent events, for which the existence probability score $P_{ex}$ is calculated $P_{ex}(A\cap\,B) = P(A) \cdot P(B|A)$.
Voxels are deemed as \textit{static} if the existence probability score $P_{ex}$ is greater than 
the static threshold probability: $P_{ex} >= P_{static}$;
otherwise, voxels represent the \textit{dynamic} state.
Points within \textit{dynamic} voxels are re-labeled to the \textit{other} class and are back-projected to the input point cloud.
The \textit{static} voxels obtain the point class that scores the greatest probability $P(B)$ within a voxel (Figure~\ref{fig:dynamicVsStatic}).

\begin{figure} [htb]
    \centering
    \includegraphics[width=0.7\linewidth]{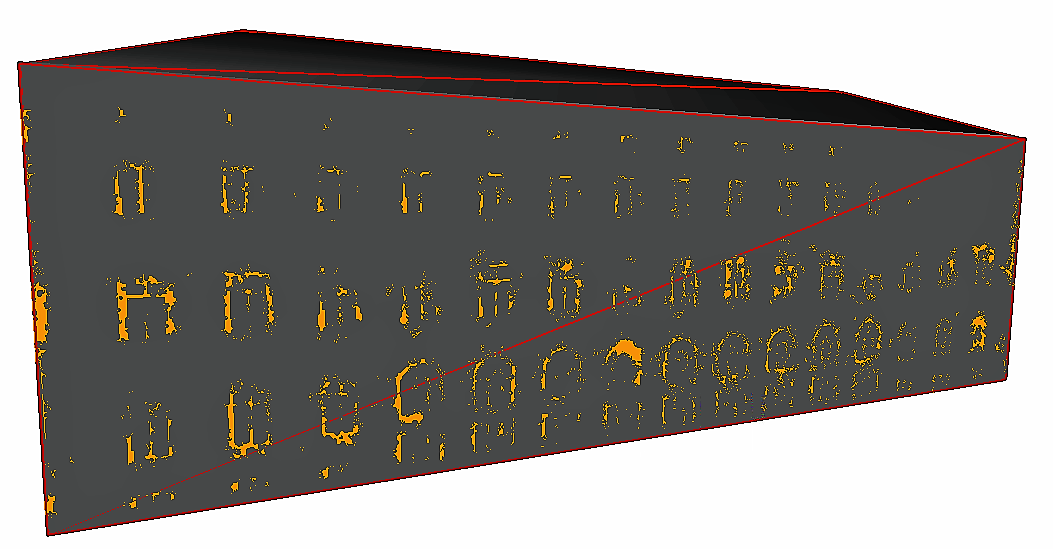}
    \caption{Texture showing one of the~\textit{static} voxel classes~\textit{window} on a façade}
    \label{fig:windowlabels}
\end{figure}

\textit{Static} voxels with semantics are projected onto the façade, forming the \textit{points comparison} texture map layer with labels corresponding to the classes, as shown in the example of windows (orange) in Figure~\ref{fig:windowlabels}.
As in the \textit{model comparison} layer (Section~\ref{sec:voxelToModel}), the cell spacing of a texture map follows the projection of the voxel grid to the plane.
\subsection{Probabilistic classification: the Bayesian approach} \label{sec:bayesian}

\textit{Model comparison} and \textit{points comparison} textures are utilized to identify façade openings using a~\gls{BN}.
The network estimations are also back-projected onto semantic point clouds to enhance their segmentation accuracy. 

\begin{figure} [htb]
    \centering
    \includegraphics[width=0.8\linewidth]{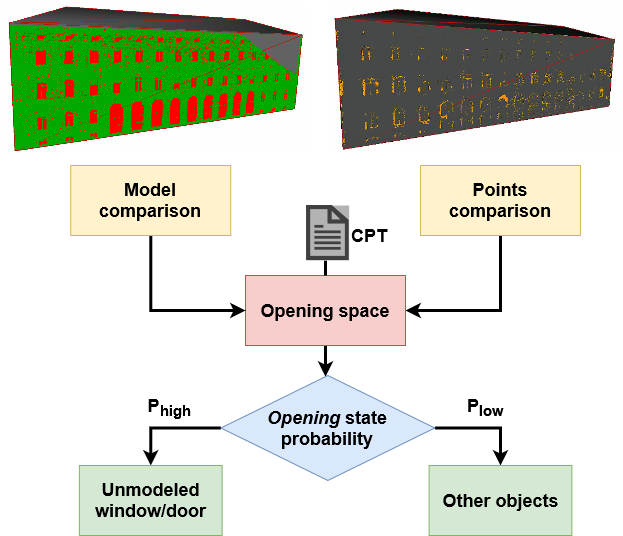}
    \caption{Input nodes (yellow) and \gls{CPT} estimate the probability of opening space (red) in \gls{BN}: if (blue) the probability is high, doors or windows are unmodeled; otherwise, areas indicate other objects (green).}
    \label{fig:bayesianNetwork}
\end{figure}

As shown in~Figure~\ref{fig:bayesianNetwork}, the designed~\gls{BN} comprises: one target (red), two input (yellow), one decision (blue), and two output nodes (green).
Each directed link represents a causal relationship between the $X$ and the $Y$ nodes.
The~\gls{CPT} prescribes weights for each state and node combination (gray).
The target, \textit{opening} state is calculated using the joint probability distribution $P(X, Y)$ and the~\gls{CPT}.
The marginalization process is used to calculate the probability of the target node $Y$ being in the \textit{opening} state $y$.
The process sums conditional probabilities of the states $x$ stemming from parent nodes $X$~\citep{stritih2020online}.
Since the network consists of texture layers with state probabilities, the data evidence represents the so-called soft evidence~\citep{stritih2020online}.
In an inference process, soft evidence is added to update the joint probability distribution.
This process provides the most likely node states by estimating the posterior probability distribution (PPD).

Pixel classes from the~\textit{model comparison} and~\textit{points comparison} textures form clusters if they have a neighbor in any of the eight directions of the pixel. 
The co-occurring~\textit{conflicted},~\textit{window}, and~\textit{door} cluster classes, lead to a high probability of unmodeled openings.
This output is used for further opening 3D modeling and is back-projected onto segmented point clouds as either the~\textit{window} or~\textit{door} class.
On the other hand, co-occurring~\textit{confirmed},~\textit{window}, and~\textit{door} clusters, lead to a low probability of existing openings.
These clusters are also back-projected to improve the accuracy of semantically segmented point clouds: 
either as the~\textit{molding} class, if close to an opening; or otherwise as the~\textit{wall} class. 
The low probability $P_{low}$ and the high probability $P_{high}$ labels are assigned to clusters based on the probability threshold $P_{t}$: $P_{high} > P_{t} >= P_{low}$.

\subsection{Openings shape extraction} \label{sec:shapeExtraction}

The high probability clusters $P_{high}$ are extracted from a Bayesian probability texture as opening shape candidates. 
Adding to existing shape indices~\citep{basaraner2017performance}, we introduce the completeness index, which measures the $r_{cp}$ ratio of outer shape area to inner-holes area.
The candidates are rejected if their area is smaller than the chosen area threshold value $b_{s}$ and if their completeness index score $r_{cp}$ is smaller than $r_{cp_{t}}$.

\subsection{Openings shape generalization} \label{sec:shapeGeneralization}

Yet, the extracted candidates can still display distorted, noisy shapes.
Morphological opening operation is applied to minimize the effect of spiky and weakly connected contours.
Subsequently, these shapes are generalized to minimum bounding boxes, for which a modified rectangularity index~\citep{basaraner2017performance} is calculated. 
The modification considers relation of the bounding box sides $a$ to $b$, where outliers are rejected based on the upper $PE_{up}$ and lower $PE_{lo}$ percentiles of the index score. 

\subsection{Model-driven 3D reconstruction} \label{sec:libraryBased}

\begin{figure} [htb]
    \centering
    \includegraphics[width=0.75\linewidth]{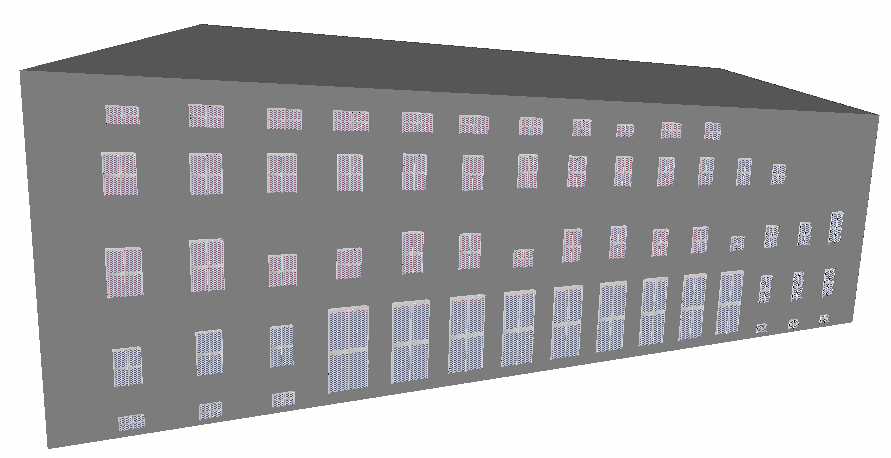}
    \caption{Reconstructed 3D windows for façade A}
    \label{fig:reconstructedWindows}
\end{figure}

\begin{figure} [htb]
    \centering
    \includegraphics[width=0.7\linewidth]{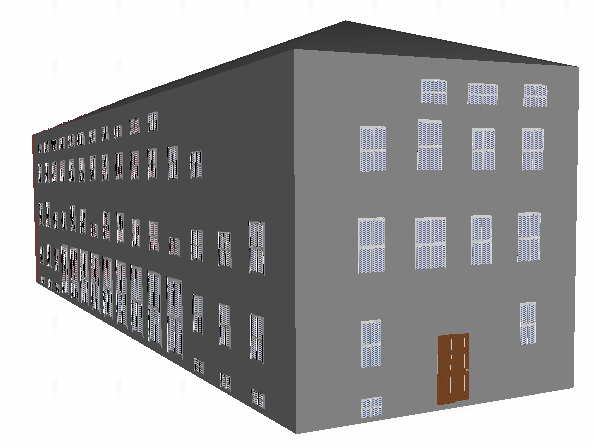}
    \caption{Reconstructed 3D windows and doors for façade B}
    \label{fig:reconstructedWindowsANDdoors}
\end{figure}

Identified bounding boxes are used as fitting boundaries for window and door 3D models, which are loaded from a pre-defined library.
The opening models' coordinate origin is erased and then placed in the bottom left corner of a model.
The offset to global coordinates is calculated between the opening model origin and the bottom left corner of the respective bounding box.
After the shift, the rotation is performed as a difference between the façade's face orientation and opening model orientation.
Aligned 3D models are scaled to fit bounding box boundaries, as presented in~Figure~\ref{fig:reconstructedWindows} and in~Figure~\ref{fig:reconstructedWindowsANDdoors}.  

\subsection{Semantic modeling} \label{sec:SemanticOpenings}

Since 3D solid libraries of openings are employed for 3D reconstruction, we opt to model them as solid geometries, too, following the CityGML encoding recommendation~\citep{special_interest_group_3d_modeling_2020}.
Based on the identified opening class, windows and doors are assigned to the respective CityGML \textit{Window} and \textit{Door} classes;
as such, they link to the building entity~\citep{grogerOGCCityGeography2012}.
The unchanged semantics of input elements is preserved, except for the \gls{LoDv2} which is upgraded to \gls{LoDv2}3. 

\section{Experiments} \label{sec:experiments}

\subsection{Datasets} \label{sec:datasets}

The method was tested using \gls{MLS} point clouds and governmental CityGML building models at \gls{LoDv2}2 representing \gls{TUM} main campus, Munich, Germany.

\begin{figure} [htb]
    \centering
    \includegraphics[width=0.7\linewidth]{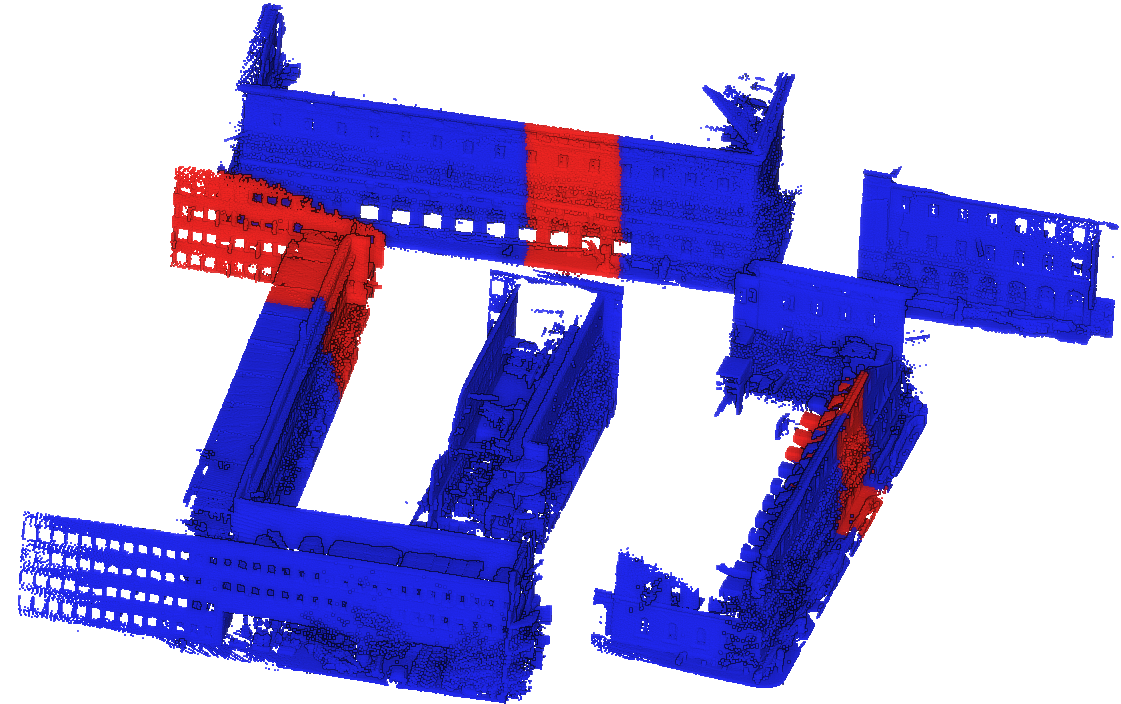}
    \caption{The TUM-FAÇADE benchmark divided into training (blue) and validation (red) sets for the segmentation experiments.}
    \label{fig:trainTest}
\end{figure}

The acquired \gls{LoDv2}2 building models were created using 2D cadastre footprints and aerial measurements~\citep{RoschlaubBatscheider}.
\gls{LoDv2}3 door and window models were extracted from the manually modeled, open \gls{LoDv2}3 city model of Ingolstadt, Germany~\footnote{https://github.com/savenow/lod3-road-space-models}.
The open TUM-MLS-2016 dataset~\citep{zhu_tum-mls-2016_2020} was transformed into the global~\gls{CRS} and used to perform point cloud ray tracing.
The TUM-FAÇADE dataset was deployed for training, as it comprises façade-annotated point clouds~\citep{tumfacadePaper}.
For computational reasons, we subsampled the original dataset removing all the redundant points within a 5 $cm$ distance. 
In this way, we compressed an initial dataset of about 118 million points to a still resolute but lightweight version of about 10 million points. 
The subsampled point cloud was divided into 70\% training and 30\% validation sets (Figure~\ref{fig:trainTest}). 
Additionally, 17 available classes were consolidated into seven representative façade classes:
\textit{molding} was merged with \textit{decoration}; \textit{wall} included \textit{drainpipe}, \textit{outer ceiling surface}, and \textit{stairs}; \textit{floor} comprised \textit{terrain} and \textit{ground surface}; \textit{other} was merged with \textit{interior} and \textit{roof}; \textit{blinds} were added to \textit{window}; while \textit{door} remained intact.

\subsection{Parameter settings} \label{sec:parameterSettings}

The uncertainties of the true façades location were estimated considering the global registration error of~\gls{MLS} point clouds and building models:
For point clouds these were set to $e_{1}= 0.3$ $m$, $\mu = 0.15$ $m$, $CL_{1}= 90\%$, and $z_{1} = 1.64$; for building models were set to $e_{2}= 0.03$ $m$, $\mu = 0.015$ $m$, $CL_{2}=90\%$, and $z_{1} = 1.64$.
This yielded the façades' upper~\gls{CI} score of 0.2 $m$ and $CL = 90\%$.

Ray casting was employed on a grid with the voxel size set to $v_{s} = 0.1$ $m$ considering: opening size, the point clouds density, and their relative accuracy.
The voxels were initialized with a uniform prior probability of $P = 0.5$.
Log-odd values were set to $l_{occ} = 0.85$ for \textit{occupied} and $l_{emp} = -0.4$ for \textit{empty} states, corresponding to $P_{occ} = 0.7$ and $P_{emp} = 0.4$, respectively.
Clamping parameters were set to $l_{min} = -2$ and $l_{max} = 3.5$, corresponding to $P_{min} = 0.12$ and $P_{max} = 0.97$, respectively, following~\citep{tuttas2015validation,hornung2013octomap};
an exemplary implementation is provided in our repository~\footnote{https://github.com/OloOcki/conflict-mls-citygml-detection}.
For the fusion of voxels and points, the static threshold was set to $P_{static} = 0.7$, while the \textit{empty} voxels occupancy probability was fixed to 0.4 for processing acceleration.

As regards the semantic segmentation procedure, taking into consideration the main characteristic of the buildings, the classes to be detected, and following \citet{grilli2019geometric}, we identified 0.8 $m$ as optimal neighborhood search radius $r_{i}$  for the features \textit{roughness, volume density, omnivariance, planarity}, and \textit{surface variation}, while 0.4 $m$ for \textit{verticality}.

The proposed~\gls{BN} has two input soft evidence layers: \textit{points comparison} and \textit{model comparison} textures.
These had associated confidence levels, which scored 70\% and 90\% for \textit{point} and \textit{model comparison} layers, respectively.
The opening state probability was defined by the probability threshold: $P_{t} = 0.7$.

The opening candidates' area threshold value $b_{s}$ was set to 0.3~$m^2$, while completeness threshold score $r_{com_{t}}$ was set to 0.1, to suppress noisy, patchy clusters.
The over-elongated bounding boxes were suppressed by calculating the modified rectangularity index, where the upper $PE_{up}$ and lower $PE_{lo}$ were set to the 95th and 5th percentile, respectively.

\subsection{Validation of improved semantic segmentation} \label{sec:semanticSegmentationExperiments}

Semantic segmentation results were validated on unseen ground-truth point clouds of the TUM-FAÇADE dataset. 
For evaluation, we use the overall accuracy (OA); F1 score per class; and average: precision ($\mu$P), recall ($\mu$R), F1 score ($\mu$F1), and intersection over union ($\mu$IoU). 
The \textit{arch} and \textit{column} classes were omitted in the validation, since they were absent in the ground-truth building. 
As shown in Table~\ref{tab:DLtesting}, for the baseline of the validation served the Point Transformer (PT) network~\citep{zhao2021point}.
The presented feature-extended version of the PT network (PT+Ft.) served as an input for the proposed conflict classification (CC) method. 

\begin{table}[h]
\setlength\tabcolsep{1pt} 
\footnotesize
	\centering
		\begin{tabular}{l|ccccc|cccccc}\hline
			 \textbf{Method} & OA & $\mu$P & $\mu$R & $\mu$F1 & $\mu$IoU & molding & floor & door & window & wall & other \\\hline
			 PT & 63.4 & 58.5 & 53.2 & 53.8 & 41.4 & 48.2 & 84.8 & 1.5 & 48.7 & 81.5 & 58.4  \\
			 PT+Ft. & 72.6 & \textbf{66.4} & 66.7 & 63.3 & 52.0 & 68.3 & 92.9 & 5.1 & 54.6 & 86.6 & 72.7\\
			 \textbf{CC} & \textbf{75.3} & 65.9 & \textbf{71.9} & \textbf{65.4} & \textbf{52.9} & 67.6 & 86.3 & 13.6 & 59.8 & 85.9 & 79.1\\\hline
		\end{tabular}
	\caption{Overall accuracy (OA), average recall ($\mu$R), average F1 ($\mu$F1), average intersection over union ($\mu$IoU), and F1 scores per class, given in percents.}
\label{tab:DLtesting}
\end{table}



\subsection{Validation of openings reconstruction} \label{sec:semanticReconstructionExperiments}

Reconstructed openings were validated using manually modeled ground-truth building openings (Table~\ref{tab:validationRecon}).
Detection rate was calculated based on the on-site inspection of all existing façade openings (AO) and measured openings (MO) by laser scanner ({Table~\ref{tab:validationDetection}}).
The validation was performed for façades A, B, and C, shown in Figure~\ref{fig:reconstructedWindows}, Figure~\ref{fig:reconstructedWindowsANDdoors}, and Figure~\ref{fig:comparisonVegetation}, respectively.
\begin{table}[h]
	\centering
		\begin{tabular}{l|ccc|c}\hline
			 Façade  & A  & B & C & \textbf{Total}\\\hline
			 AO & 66 & 17 & 20 & \textbf{101}  \\
			 MO & 60 & 17 & 10 & \textbf{87}  \\
			 D & 60 & 15 & 6 & \textbf{81} \\
			 TP & 60 & 15 & 5 & \textbf{80} \\
			 FP & 0 & 0 & 1 & \textbf{1} \\
			 FN & 6 & 2 & 12 & \textbf{21} \\\hline
			 \textbf{DR-AO} & 90.1\% & 88.2\% & 27.8\% & \textbf{79.2\%}  \\
			 \textbf{FR-AO} & 0\% & 0\% & 16.7\% & \textbf{1.2\%}  \\
			 \textbf{DR-MO} & 100\% & 88.2\% & 50\% & \textbf{91.9\%}  \\
			 \textbf{FR-MO} & 0\% & 0\% & 16.7\% & \textbf{1.1\%}  \\\hline
		\end{tabular}
	\caption{Detection rate for all openings (DR-AO) and measured openings (DR-MO) and the respective false alarm rate (FR-AO and FR-MO) for façades A, B, and C (D = detections, TP = true positives, FP = false positives, FN = false negatives, for all openings). }
\label{tab:validationDetection}
\end{table}

\begin{table}[h]
	\centering
		\begin{tabular}{l|ccc|c}\hline
			 Façade  & A  & B & C & \textbf{Total}\\\hline
			 $m$IoU & 88.4\%  & 94.8\% & 97.2\% & \textbf{89.6\%}\\
			 $\mu$IoU & 79.1\% & 83.4\% & 85.7\% & \textbf{80.3\%}\\\hline
		\end{tabular}
	\caption{Validation of reconstructed openings using median ($m$IoU) and average intersection over union ($\mu$IoU).}
\label{tab:validationRecon}
\end{table}



\begin{figure*}
    \centering
    \includegraphics[width=0.8\linewidth]{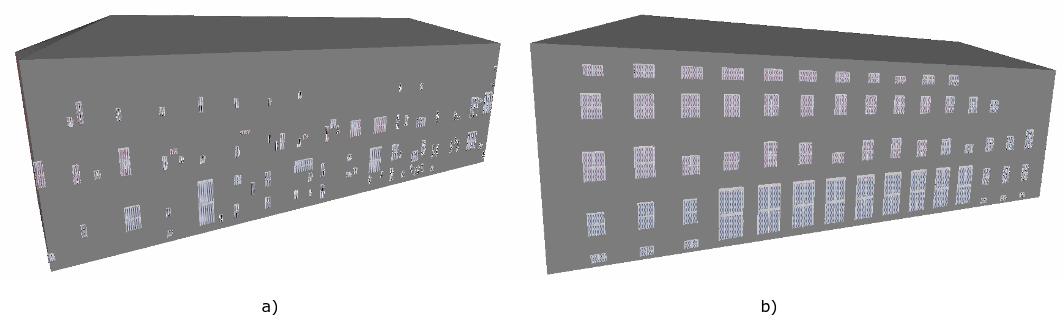}
    \caption{Windows reconstructed using: a) only neural networks output, b) the proposed method.}
    \label{fig:comparisonReconstructedWindows}
\end{figure*}

\begin{figure*}
    \centering
    \includegraphics[width=0.8\linewidth]{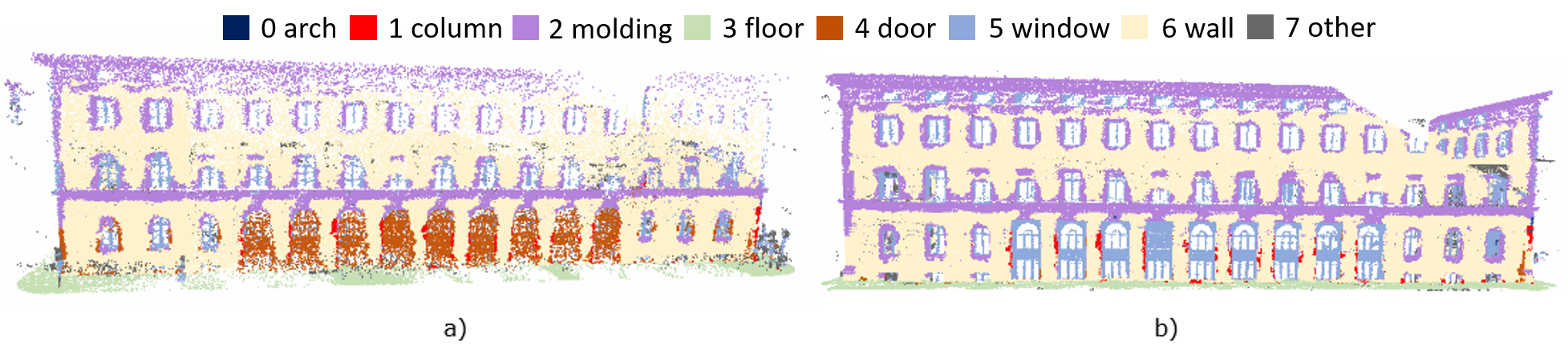}
    \caption{Semantic segmentation based on: a) only a neural network, b) the proposed method.}
    \label{fig:comparisonPointClouds}
\end{figure*}

\section{Discussion} \label{sec:discuss}
\begin{figure}
    \centering
    \includegraphics[width=\linewidth]{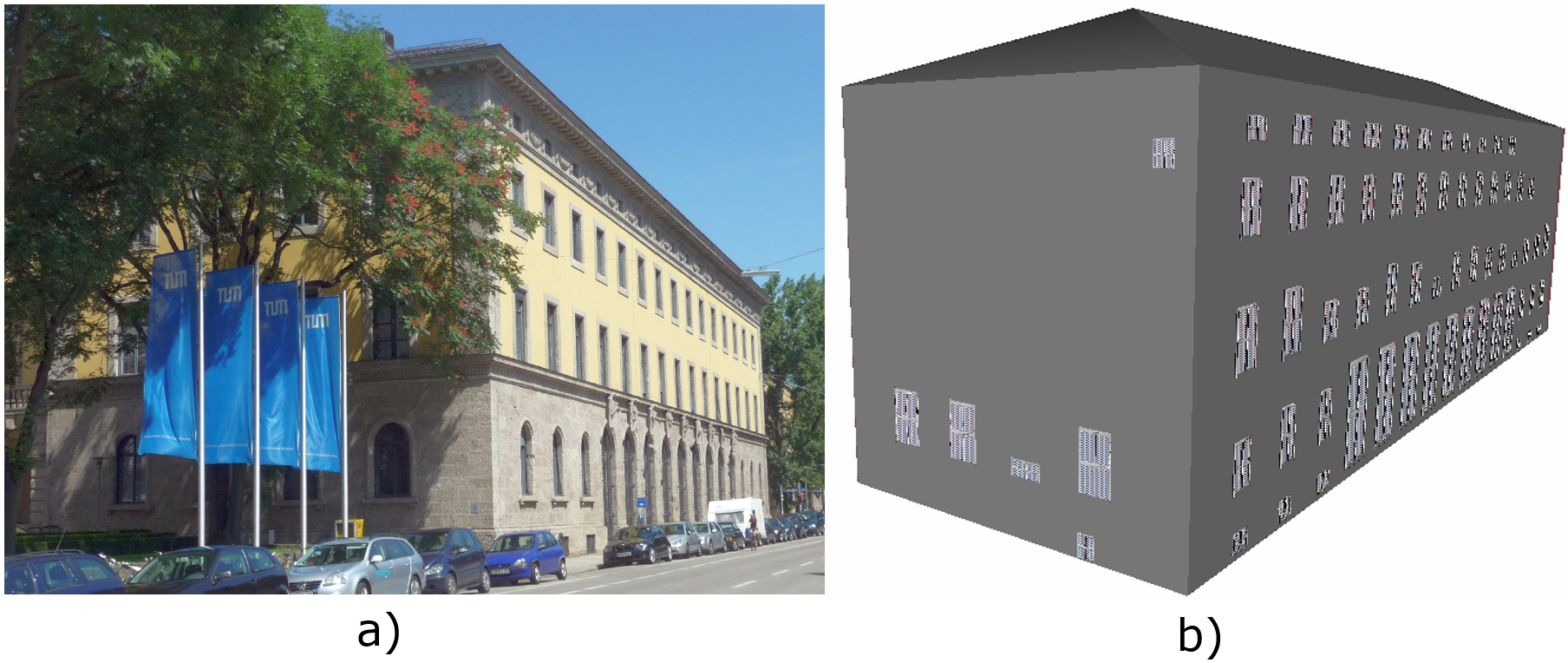}
    \caption{Openings reconstruction for façade C in the presence of occluding objects: a) Photo, b) refined façade.}
    \label{fig:comparisonVegetation}
\end{figure}
Experiments revealed promising results for refining of both building models and classified point clouds.
As presented in Table~\ref{tab:validationDetection}, façade openings were correctly detected with an estimated 92\% detection rate to total measured openings (DR-MO) and 79\% detection rate for all openings (DR-AO).
Roughly 1\% false alarm rate for both measured (FR-MO) and all openings (FR-AO) was noted.

The experiments corroborate that DR was dependent on the density of measurements per façade: for the densely covered façade A it estimated 90\% DR-AO and detected 100\% of measured openings (DR-MO); for the highly occluded side-façade C it estimated 28\% and 50\%, respectively (see Table~\ref{tab:validationDetection} and Figure~\ref{fig:comparisonVegetation}).

The method improves significantly reconstruction performance in comparison to the one conducted only on segmented point clouds of the baseline PT architecture, as shown in Figure~\ref{fig:comparisonReconstructedWindows}.
When compared to the ground-truth openings, the proposed reconstruction reached roughly 90\% accuracy (Table~\ref{tab:validationRecon});
yet, the method is limited when windows are partially measured (e.g., blinds before windows), as exemplified by several windows in the third row in Figure~\ref{fig:comparisonReconstructedWindows}b.

The back-projected, classified conflicts increased accuracy of semantic point cloud segmentation by approximately 12\% (Table~\ref{tab:DLtesting}).
Note that the precision and intersection over union score for CC remained similar to the PT+Ft. score, while F1 score for~\textit{floor} dropped by about 6\%.  
Remarkably, the proposed CC method improves segmentation of~\textit{window},~\textit{door}, and~\textit{other} classes by approximately 11\%, 12\%, and 21\%, respectively.

\section{Conclusion} \label{sec:conclusion}

Our work has led us to the conclusion that refinement is a promising alternative to a from-scratch reconstruction.
The refinement preserves input semantics, minimizes model-specific planarity issues, and enables consistent city model updates. 
Moreover, existing~\gls{LoDv2}3 elements can be extracted and directly employed as refinement features for buildings at lower~\gls{LoDv2}s.

The validation presents that the method reaches a high accuracy of 92\% in detecting observable windows and a low false alarm rate score of approximately 1\%.
Refined point clouds also score low false negative rate, which is indicated by a high recall score of 79\%.
This trait of our method could be of particular importance for feature-dependent applications, where robustness is favored over visualization, such as in simulations of automated driving functions~\citep{schwabRequirementAnalysis3d2019}.
On the other hand, façade occlusions and a laser range could limit the method's applicability for visualization-oriented purposes, where a further prediction of unseen objects could be employed.

Experiments corroborate that combining visibility analysis with a region-based approach improves segmentation accuracy. 
In the future, we plan to embed occupancy information directly into the training of the deep neural network.  
Furthermore, including radial point cloud features (e.g., intensity) in training datasets could facilitate detecting windows covered by blinds.

Tested façades presented challenging, varying measuring conditions;
for similar façade and opening styles, the method is expected to provide comparable results.
Yet, testing sample size implies that caution must be exercised.
It is worth noting that~\textit{static} objects, which do not contribute to façades elements and are adjacent (e.g., traffic signs, bus shelters), can negatively influence the semantic back-projection results.  
To further our research, we plan to test the method on a higher number of façades.

\section*{Acknowledgments} \label{sec:acknowledgments}

This work was supported by the Bavarian State Ministry for Economic Affairs, Regional Development and Energy within the framework of the IuK Bayern project \textit{MoFa3D - Mobile Erfassung von Fassaden mittels 3D Punktwolken}, Grant No.\ IUK643/001.
Moreover, the work was conducted within the framework of the {Leonhard Obermeyer Center} at the Technical University of Munich (TUM).
We gratefully acknowledge the Geoinformatics team at the TUM for the valuable insights and for providing the CityGML datasets.

{
	\begin{spacing}{0.0}
		\footnotesize
		\bibliography{ISPRSguidelines_authors.bib} 
	\end{spacing}
}

\end{document}